\documentclass[preprint,12pt]{elsarticle}

\usepackage{amsmath,amsfonts}
\usepackage{algorithmic}
\usepackage{array}
\usepackage[tight,footnotesize]{subfigure}
\usepackage{textcomp}
\usepackage{stfloats}
\usepackage{url}
\usepackage{verbatim}
\usepackage{graphicx}
\def\BibTeX{{\rm B\kern-.05em{\sc i\kern-.025em b}\kern-.08em
    T\kern-.1667em\lower.7ex\hbox{E}\kern-.125emX}}
\usepackage{balance}
\usepackage{booktabs}
\usepackage{caption}
\usepackage{makecell}
\usepackage{adjustbox}
\usepackage{wrapfig}
\usepackage[linesnumbered,ruled]{algorithm2e}
\usepackage{amssymb}
\usepackage{lineno}
\usepackage{float}
\usepackage{algorithmic}
\usepackage[tight,footnotesize]{subfigure}
\usepackage{booktabs}
\usepackage{caption}
\usepackage{makecell}
\usepackage{adjustbox}
\usepackage{wrapfig}
\usepackage{xcolor}
\usepackage[linesnumbered,ruled]{algorithm2e}

\journal{Journal Name}

\begin{document}
\sloppy
\setlength{\parskip}{0pt}

\begin{frontmatter}




\title{Flowr --- Scaling Up Retail Supply Chain Operations Through Agentic AI in Large Scale Supermarket Chains}


\author[label1]{Eranga Bandara}
\ead{cmedawer@odu.edu}

\author[label1]{Ross Gore}
\ead{rgore@odu.edu}

\author[label1]{Sachin Shetty}
\ead{sshetty@odu.edu}

\author[label10]{Piumi Siyambalapitiya}
\ead{dpsiyambalapitiya@gmail.com}

\author[label7]{Sachini Rajapakse}
\ead{sachini.rajapakse@iciclelabs.ai}

\author[label7]{Isurunima Kularathna}
\ead{isurunima.kularathna@iciclelabs.ai}

\author[label10]{Pramoda Karunarathna}
\ead{pramodabhavani@sjp.ac.lk}

\author[label1]{Ravi Mukkamala}
\ead{mukka@odu.edu}

\author[label1]{Peter Foytik}
\ead{pfoytik@odu.edu}

\author[label1]{Safdar H. Bouk}
\ead{sbouk@odu.edu}

\author[label3]{Abdul Rahman}
\ead{abdulrahman@deloitte.com}

\author[label4]{Xueping Liang}
\ead{xuliang@fiu.edu}

\author[label8]{Amin Hass}
\ead{amin.hassanzadeh@accenture.com}

\author[label2]{Tharaka Hewa}
\ead{tharaka.hewa@oulu.fi}

\author[label5]{Ng Wee Keong}
\ead{awkng@ntu.edu.sg}

\author[label6]{Kasun De Zoysa}
\ead{kasun@ucsc.cmb.ac.lk}

\author[label9]{Aruna Withanage}
\ead{aruna@effectz.ai}
\author[label9]{Nilaan Loganathan}
\ead{nilaan@effectz.ai}

\address[label1]{Old Dominion University, Norfolk, VA, USA}
\address[label3]{Deloitte \& Touche LLP, USA}
\address[label4]{Florida International University, USA}
\address[label5]{Nanyang Technological University, Singapore}
\address[label6]{University of Colombo, Sri Lanka}
\address[label2]{Center for Wireless Communications, University of Oulu, Finland}
\address[label7]{IcicleLabs.AI}
\address[label10]{University of Sri Jayewardenepura, Sri Lanka}
\address[label8]{Accenture Technology Labs, Arlington, VA, USA}
\address[label9]{Effectz.AI}

\begin{abstract}

Retail supply chain operations in large supermarket chains involve continuous, high-volume manual workflows spanning demand forecasting, procurement, supplier coordination, and inventory replenishment — processes that are inherently repetitive, decision-intensive, and difficult to scale without significant human effort. Despite growing investment in data analytics, the decision-making and coordination layers of these workflows remain predominantly manual, reactive, and fragmented across outlets, distribution centers, and supplier networks. This paper introduces a novel agentic AI framework, Flowr, for automating end-to-end retail supply chain workflows in large-scale supermarket operations. The proposed framework systematically decomposes manual supply chain operations into a set of specialized AI agents, each responsible for a clearly defined cognitive role within the workflow, enabling end-to-end automation of processes that were previously dependent on continuous human coordination. To ensure task accuracy, contextual reasoning, and adherence to responsible and explainable AI principles, the framework employs a consortium of fine-tuned, domain-specialized large language models coordinated by a central reasoning LLM, enabling agents to reason effectively within the specific operational context of retail supply chain management. Central to the framework is a human-in-the-loop orchestration model in which supply chain managers supervise, validate, and intervene across workflow stages via a Model Context Protocol (MCP)–enabled interface, preserving accountability, transparency, and organizational control. Experimental evaluation demonstrates that the proposed agentic workflow significantly reduces manual coordination overhead, improves demand-supply alignment across store locations, and enables proactive exception handling at a scale unachievable through manual processes. As a proof of concept, the framework was implemented and validated in collaboration with a large-scale supermarket chain. Although the proof of concept is grounded in a real-world retail grocery operational environment, the proposed framework is domain-independent and offers a generalizable blueprint for agentic AI–driven supply chain automation across large-scale enterprise settings.


\end{abstract}

\begin{keyword}
Agentic AI \sep Agentic AI Workflow \sep Responsible AI \sep Explainable AI \sep LLM \sep Model Context Protocol
\end{keyword}

\end{frontmatter}

\section{Introduction}

Retail supply chain management is one of the most operationally complex and coordination-intensive functions within large-scale supermarket enterprises. Managing the continuous flow of goods across hundreds of outlets, multiple distribution centers, and thousands of suppliers requires sustained human effort across demand forecasting, procurement, inventory monitoring, supplier negotiation, and replenishment planning~\cite{jannelli2024}. These workflows are inherently repetitive, time-sensitive, and dependent on the synthesis of large volumes of data from heterogeneous sources, including point-of-sale systems, inventory databases, supplier communication channels, and external signals such as seasonal demand patterns and market disruptions. Despite significant investment in enterprise resource planning systems and data analytics platforms, the decision-making and coordination layers of these workflows remain predominantly manual in most large retail organizations, creating persistent inefficiencies, delayed responses to demand changes, and limited scalability as outlet networks expand~\cite{brintrup2026}.

The emergence of agentic AI systems represents a fundamental shift in how such operational complexity can be addressed~\cite{agentic-ai, agent-survey}. Unlike conventional AI-assisted tools that support individual decisions in isolation, agentic AI systems are capable of autonomous reasoning, multi-step planning, and coordinated action across interconnected workflows without continuous human intervention~\cite{agentic-ai-opptunities}. Networks of specialized AI agents can continuously monitor inventory levels, interpret demand signals, generate procurement orders, coordinate with suppliers, and plan distribution center replenishment runs — tasks that currently consume substantial human effort and are prone to coordination errors when performed manually~\cite{agentic-ai-rise}. By delegating these cognitive responsibilities to autonomous agents, operating under human supervision, organizations can achieve levels of operational efficiencies, throughput, responsiveness, and consistency that manual processes cannot sustain at scale.

Despite this potential, the application of agentic AI to retail supply chain management remains largely unexplored in practice. Most retail AI deployments focus on isolated capabilities such as demand prediction models, pricing optimization tools, or customer recommendation engines, rather than end-to-end workflow automation~\cite{agentic-workflow-practicle-guide}. These point solutions improve specific metrics but do not fundamentally transform the coordination structure of supply chain operations. The integration of multiple autonomous agents into a cohesive, human-supervised workflow — one capable of executing the full replenishment lifecycle from demand sensing to supplier confirmation and distribution center dispatch — represents an open challenge that existing approaches have not adequately addressed~\cite{agentic-ai-transition-organization}.

A central barrier to this transition is the absence of a principled framework for decomposing complex retail supply chain workflows into agent-executable tasks, designing the coordination interfaces between agents, and maintaining meaningful human oversight throughout the process~\cite{agentic-ai-taxonomy-challenges}. Traditional software engineering approaches are poorly suited to this challenge because agentic AI workflows are inherently probabilistic, adaptive, and context-dependent rather than deterministic and rule-bound. At the same time, retail supply chains involve high-stakes operational decisions — including procurement commitments, supplier relationship management, and perishable inventory control — where errors have direct financial and service quality consequences. This makes responsible AI design, explainability, and human-in-the-loop oversight essential requirements rather than optional enhancements~\cite{proof-of-tbi}.

In response to these challenges, this paper proposes Flowr, a novel agentic AI framework for automating end-to-end retail supply chain workflows in large-scale supermarket operations. Flowr introduces a multi-agent architecture in which manual supply chain processes are systematically decomposed into specialized AI agents, each responsible for a clearly defined cognitive role within the workflow~\cite{agentic-ai-workflow-patterns}. Agents are coordinated through a human-in-the-loop orchestration model in which supply chain managers supervise, validate, and intervene at defined checkpoints via a Model Context Protocol (MCP)–enabled interface~\cite{mcc, mcp1,  mcp2}. To ensure contextual accuracy and responsible and explainable decision-making, the framework employs a consortium of fine-tuned, domain-specialized large language models (LLMs) coordinated by a central reasoning LLM, enabling agents to reason effectively within the specific operational realities of retail supply chain management. Rather than replacing human expertise, Flowr shifts human effort from manual execution and coordination toward oversight, exception handling, and strategic decision-making. This enables scalable, trustworthy, and adaptive supply chain automation.

As a proof of concept, Flowr was implemented and evaluated in collaboration with a large-scale supermarket chain. The operational environment exhibits characteristics that are highly amenable to agentic AI automation, including high transaction volumes, diverse supplier networks, perishable inventory constraints, and geographically distributed outlets with location-specific demand patterns. These characteristics render it a representative and rigorously challenging validation environment for the proposed framework. The primary contributions of this paper are fourfold:


\begin{enumerate}
    \item A novel agentic AI framework, Flowr, for end-to-end automation of retail supply chain workflows, encompassing demand forecasting, inventory monitoring, procurement, supplier coordination, distribution center replenishment planning, and exception handling under a unified multi-agent architecture.


    \item A human-in-the-loop orchestration model in which supply chain managers act as supervisors and decision-makers over autonomous agent workflows via MCP-enabled interfaces, ensuring accountability, transparency, and operational control throughout the replenishment lifecycle.

    \item A responsible and explainable AI design incorporating a consortium of fine-tuned, domain-specialized LLMs coordinated by a reasoning LLM, supporting explainable, context-aware, and ethically aligned decision-making across all agent interactions.

    \item A real-world proof of concept demonstrating the practical applicability of Flowr within a large-scale supermarket operational environment, validating the framework's effectiveness in reducing manual coordination overhead, improving demand-supply alignment across distributed outlet networks, and enabling proactive exception handling at enterprise scale.
\end{enumerate}

The remainder of this paper is organized as follows. Section 2 provides background on key concepts underlying the Flowr framework, including agentic AI systems, multi-agent workflow design, LLM consortiums, and human-in-the-loop orchestration. Section 3 describes the manual supply chain workflows that Flowr targets and establishes the operational context. Section 4 presents the Flowr framework, detailing the multi-agent architecture, agent roles, coordination model, and LLM consortium design. Section 5 describes the proof-of-concept implementation of Flowr within a real-world, large-scale supermarket operational environment. Section 6 presents the evaluation results and findings. Section 7 reviews related work in agentic AI, retail supply chain automation, and LLM-based workflow systems. Section 8 concludes the paper and outlines directions for future research.

\section{Background}

This section introduces the key concepts and technologies that underpin the Flowr framework. Understanding these foundations is necessary before examining how they are applied to automate retail supply chain workflows in subsequent sections.

\subsection{Large Language Models}

Large Language Models (LLMs) are a class of artificial intelligence systems trained on vast corpora of text data, enabling them to understand, generate, and reason over natural language with a high degree of fluency and contextual awareness~\cite{llm}. Unlike earlier generation AI models that were trained for narrow, task-specific purposes, LLMs acquire broad general knowledge and linguistic capability through exposure to diverse text sources, allowing them to perform a wide range of tasks — including summarisation, question answering, structured data extraction, and complex reasoning — through natural language interaction alone~\cite{llm-security}.

In the context of supply chain automation, LLMs serve as the reasoning engine behind each specialized agent in the Flowr framework. Their ability to interpret unstructured inputs — such as supplier communication messages, inventory exception reports, and demand signals — and produce structured, actionable outputs makes them particularly well-suited to the judgment-intensive tasks that characterise retail supply chain operations. Rather than relying on rigid rule-based logic, LLM-powered agents can reason flexibly over incomplete information, adapt to changing operational conditions, and generate outputs that are interpretable by human supervisors~\cite{llm-agents}.

\subsection{Reasoning LLMs}

While standard LLMs excel at language understanding and generation, reasoning LLMs extend these capabilities by incorporating explicit chain-of-thought processes, multi-step planning, and self-verification mechanisms that enable more reliable and structured decision-making~\cite{reasoning-llms, nurolense}. Reasoning LLMs are designed to evaluate intermediate conclusions, identify inconsistencies across multiple sources of evidence, and synthesise coherent final outputs from complex, multi-perspective inputs.

In the Flowr framework, a central reasoning LLM — specifically OpenAI GPT-OSS~\cite{gpt-oss} — serves as the supervisory model within the LLM consortium architecture. When multiple fine-tuned domain-specialized LLMs produce independent outputs in response to an agent's prompt, the reasoning LLM evaluates, compares, and synthesises these responses to generate a final decision or recommendation. This ensemble-based reasoning approach reduces the risk of model-specific bias or error propagation, and ensures that automated supply chain decisions are validated across multiple reasoning perspectives before execution — a critical requirement in operational environments where procurement errors have direct financial consequences~\cite{llm-reasoning}.

\subsection{LLM Fine-Tuning}

While pre-trained LLMs possess broad general knowledge, their performance on domain-specific tasks — particularly those requiring precise operational vocabulary, structured output formats, and context-aware reasoning within a specific industry — can be significantly improved through fine-tuning~\cite{mistral-fine-tune, proof-of-tbi}. Fine-tuning is the process of adapting a general-purpose pre-trained base model on a curated, domain-specific dataset through supervised training, enabling the model to internalise domain conventions, terminologies, and reasoning patterns that are not well-represented in general pre-training corpora.

\begin{figure}[H]
\centering{}
\includegraphics[width=5.4in]{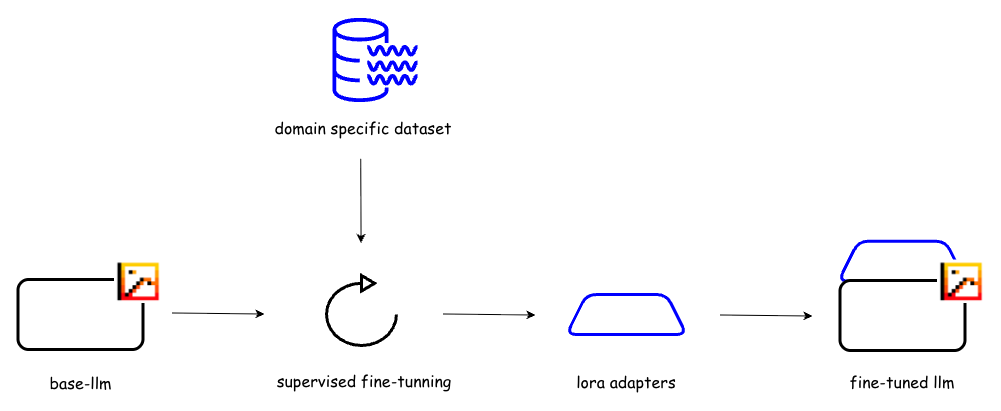}
\DeclareGraphicsExtensions.
\caption{Overview of the supervised fine-tuning pipeline used to adapt a base LLM to domain-specific tasks. A curated, domain-specific dataset is used to perform supervised fine-tuning of a base LLM, with Low-Rank Adapter (LoRA) modules applied to enable parameter-efficient training. The resulting fine-tuned model retains the general linguistic capabilities of the base model while incorporating domain-specific knowledge, supporting efficient deployment and updates in resource-constrained and air-gapped environments.}
\label{llm-fine-tune}
\end{figure}

As illustrated in Figure~\ref{llm-fine-tune}, the fine-tuning pipeline employed in the Flowr framework begins with a base LLM — such as Llama-3, Mistral, or Qwen — which is adapted using a curated retail supply chain dataset comprising historical sales records, supplier interaction logs, procurement order histories, and outlet-level inventory records. To enable parameter-efficient training within resource-constrained environments, Low-Rank Adapters (LoRA)~\cite{lora} are applied during the fine-tuning process, and 4-bit quantization (QLoRA)~\cite{qlora} is employed to further reduce memory requirements. Fine-tuned models are deployed locally using Ollama~\cite{ollama}, enabling low-latency agent inference without dependence on external API services. The resulting fine-tuned models retain the general linguistic capabilities of their base counterparts while incorporating the domain-specific knowledge necessary to reason accurately over retail supply chain tasks — from demand interpretation and supplier selection reasoning to replenishment planning and exception detection.

\subsection{AI Agents and Agentic AI}

Traditional LLM interactions follow a simple request--response pattern in which a human constructs a prompt, submits it to a language model, and manually interprets the response to determine subsequent actions. In this paradigm, the human remains responsible for orchestration, decision-making, and follow-up execution, with the LLM functioning as a passive assistant rather than an active participant in workflow execution~\cite{agentic-ai, agentic-workflow-practicle-guide}.

\begin{figure}[H]
\centering
\includegraphics[width=5.3in]{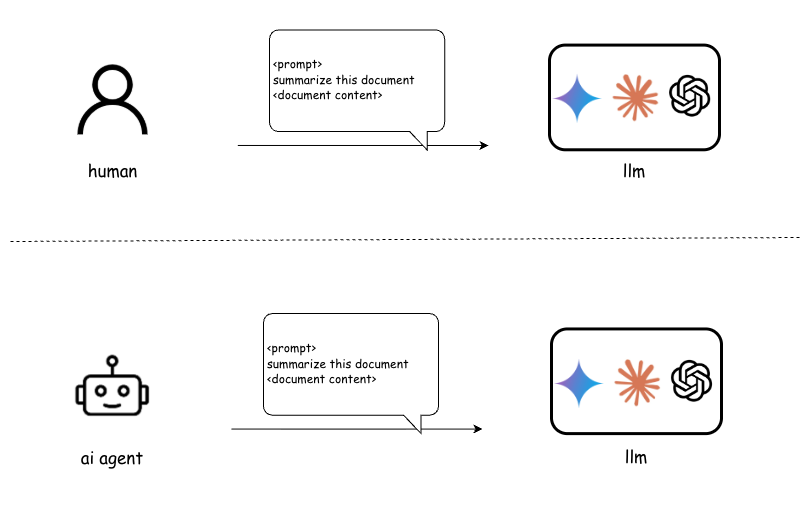}
\vspace{-0.1in}
\caption{Comparison between direct human--LLM interaction and agentic AI--LLM interaction. In the top panel, a human user directly constructs and submits a prompt to a language model, manually interpreting the response. In the bottom panel, an AI agent autonomously manages the interaction loop by constructing prompts, invoking the language model, interpreting responses, and determining subsequent actions without direct human intervention. This agent-mediated workflow enables iterative reasoning, tool integration, and task automation, forming the basis of agentic AI systems used for structured decision support and multi-step workflows.}
\label{ai-agent}
\end{figure}

An AI agent fundamentally changes this interaction model. As illustrated in Figure~\ref{ai-agent}, an AI agent autonomously manages the full interaction loop — constructing prompts, invoking language models, interpreting responses, and determining subsequent actions — without direct human intervention at each step. AI agents are software systems that leverage LLMs in combination with tools, APIs, memory, and external context to execute tasks iteratively and autonomously~\cite{agentsway}. When multiple specialized agents collaborate — each assigned distinct responsibilities such as data retrieval, content filtering, reasoning, validation, or publication — they form \textit{agentic AI workflows}~\cite{agentic-ai-transition-organization}.

Agentic AI workflows enable systems that can reason over complex, multi-step tasks, plan sequences of actions, interact with external systems, monitor outcomes, and adapt their behavior through iterative feedback. In the Flowr framework, this agentic paradigm is applied to retail supply chain management — with each manual process stage delegated to a specialized AI agent capable of autonomous reasoning and action within its defined scope, while the overall workflow remains coordinated and supervised by a human orchestrator. This shift from tool-centric AI assistance to workflow-centric agentic automation is central to the operational value that Flowr delivers.

\subsection{Model Context Protocol}

The Model Context Protocol (MCP) is a standardized, open protocol that defines how AI agents connect to and interact with external systems, tools, databases, and APIs through secure, structured interfaces~\cite{mcp1, mcp2, mcc}. MCP enables AI agents to access real-time data and invoke external capabilities — such as querying inventory databases, retrieving supplier catalogues, or submitting purchase orders — in a modular and interoperable manner, without requiring custom integration code for each system connection.

In the Flowr framework, each agentic workflow is exposed as an independent MCP server, enabling a single human supply chain manager to connect to and orchestrate multiple agent workflows simultaneously through a unified natural language interface via LM Studio~\cite{lm-studio}. This architecture decouples the agent logic from the underlying systems it interacts with — the Inventory Monitoring Agent connects to the inventory management system through its MCP server, the Supplier Coordination Agent communicates with supplier systems through its own MCP server, and so on — allowing individual workflows to evolve independently without disrupting the broader system. MCP also provides a natural human coordination layer: the supply chain manager invokes workflows, reviews outputs, and approves actions through the same MCP-enabled interface, maintaining full visibility and control over the automated pipeline.

\subsection{Responsible and Explainable AI}

Responsible and Explainable AI principles emphasize transparency, accountability, fairness, and human oversight in AI system design. In high-stakes domains such as military operations and mental health support, these principles are critical for ensuring trust, safety, and ethical deployment~\cite{towards-rai-xai}.

Explainable AI focuses on providing interpretable outputs and rationales that allow human operators to understand and evaluate system recommendations. Responsible AI further requires clear operational boundaries, avoidance of autonomous decision-making in sensitive contexts, and mechanisms for auditability and oversight~\cite{responsible-gen-ai, llm-explainability}.

The proposed framework operationalizes these principles through strict human-in-the-loop control, consensus-driven reasoning across multiple models, structured logging of agent interactions, and offline learning mechanisms. This design ensures that agentic AI augments human capability without displacing human judgment or authority.

\subsection{Responsible and Explainable AI}

As agentic AI systems take on increasingly consequential operational roles — generating procurement orders, selecting suppliers, and planning distribution schedules — ensuring that their decisions are transparent, accountable, and aligned with organizational and ethical principles becomes a foundational requirement rather than an optional enhancement~\cite{responsible-ai, xai}. Responsible AI encompasses the principles of fairness, transparency, accountability, privacy, and human oversight in the design and deployment of AI systems. Explainable AI (XAI) refers specifically to the ability of AI systems to provide human-interpretable justifications for their outputs — enabling operators to understand not just what a system decided, but why~\cite{deep-psychiatric}.

In the Flowr framework, responsible and explainable AI is operationalized through three complementary mechanisms. First, the LLM consortium architecture ensures that all automated decisions are validated across multiple independent reasoning perspectives before execution, reducing the risk of model-specific bias or hallucination in high-stakes supply chain decisions. Second, all agent outputs are structured to include explicit reasoning traces — every purchase order includes a supplier selection justification, every replenishment plan includes an allocation rationale, and every exception alert includes the specific threshold or condition that triggered it — enabling human supervisors to audit and verify agent behavior at any point in the workflow. Third, human approval gates are embedded at critical decision points throughout the pipeline, ensuring that no consequential action — such as confirming a supplier order or dispatching a replenishment plan — is executed without explicit human validation. Together, these mechanisms ensure that the Flowr framework delivers operational automation while preserving the accountability, trust, and organizational control that responsible AI deployment requires.

\section{Manual Supply Chain Workflows in Large Supermarket Operations}

\begin{figure}[H]
\centering
\includegraphics[width=5.4in]{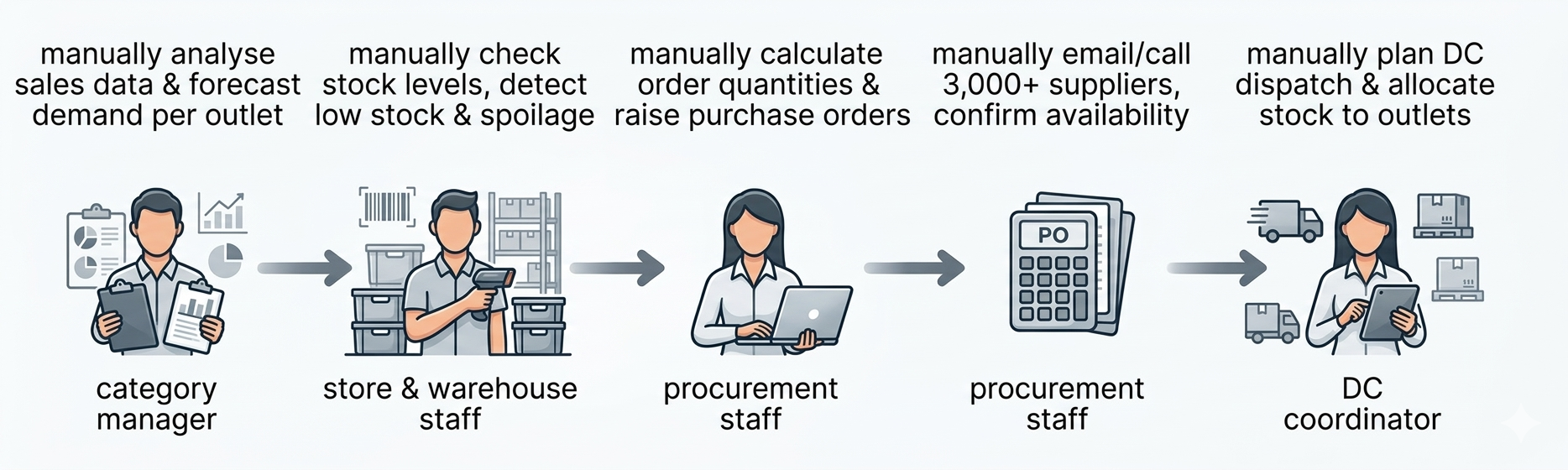}
\vspace{-0.1in}
\caption{Manual supply chain replenishment workflow in large supermarket operations. Category managers, store and warehouse staff, procurement teams, and distribution center coordinators execute sequential stages — demand analysis, inventory monitoring, purchase order generation, supplier coordination, and outlet replenishment planning — across disconnected systems with no automated integration layer. Exceptions such as stockouts, supplier delays, spoilage risks, and demand spikes are detected reactively and escalated manually, limiting the organization's ability to respond proactively at scale.}
\label{manual-flow}
\end{figure}

A successful agentic AI transition begins with a thorough understanding of the manual processes that automation intends to replace. In the context of large-scale supermarket operations, the supply chain function encompasses a set of recurring, coordination-heavy workflows that span demand sensing, procurement, supplier communication, inventory management, and physical replenishment across geographically distributed outlet networks~\cite{brintrup2026}. These workflows are deeply interdependent — the output of one stage directly conditions the inputs of the next — yet in practice, they are executed by different teams, across different systems, with limited real-time integration. As depicted in Figure~\ref{manual-flow}, the end-to-end replenishment lifecycle is carried out sequentially by distinct human roles — from category managers performing demand analysis through to distribution center coordinators planning outlet dispatch — with no automated coordination layer connecting these stages. The workflows are highly repetitive, involve continuous synthesis of information from multiple heterogeneous sources, require sustained coordination across large supplier and outlet networks, and are sensitive to timing constraints that manual processes frequently cannot satisfy at scale. These characteristics — continuous monitoring requirements, repeated decision-making, cross-source reasoning, and the need for human oversight at critical decision points — are precisely the conditions under which agentic AI workflows deliver the greatest operational value, as identified in our earlier work on agentic AI transition frameworks~\cite{junqueira2011zab}. The manual processes underlying each stage of this workflow are described in detail in the following subsections.

\subsection{Demand Forecasting and Sales Analysis}

The demand forecasting process begins when category managers and procurement staff manually extract sales data from point-of-sale systems on a periodic basis, typically weekly or monthly. This data is reviewed alongside historical records, seasonal calendars, and promotional schedules to produce informal demand estimates for individual product categories~\cite{ren2025}. In most large supermarket operations, this analysis is performed using spreadsheets or basic business intelligence dashboards, requiring significant manual effort to consolidate data across multiple outlets and product lines. Demand indicators derived from exogenous sources—such as forthcoming public holidays, meteorological conditions, or competitor promotional activities—are currently integrated in an ad hoc manner based on individual analyst discretion, rather than through a standardized, systematic data integration framework. The resulting forecasts are manually communicated to procurement teams, often through email or internal messaging systems, introducing delays and version control issues that compound downstream.

\subsection{Inventory Monitoring and Stockout (Out-of-Stock) Detection}

Inventory monitoring is performed by store-level staff and warehouse teams who manually check stock levels against minimum threshold records at regular intervals. In larger outlet networks, this process is partially supported by inventory management systems that generate low-stock alerts, but the interpretation of these alerts, the prioritization of replenishment actions, and the identification of spoilage risks for perishable categories remain manual responsibilities~\cite{ren2025marl}. Category managers review inventory reports periodically and cross-reference them with demand forecasts to identify replenishment needs, a process that is time-consuming and reactive by nature. Discrepancies between system records and physical stock counts — caused by shrinkage, misallocation, or data entry errors — further complicate the monitoring process and require manual investigation and correction. Perishable inventory management is particularly demanding, as staff must continuously assess product freshness, anticipate expiry risks across multiple outlets, and coordinate urgent replenishment actions before stock loss occurs.

\subsection{Procurement and Purchase Order Generation}

Once replenishment needs are identified, procurement staff manually calculate order quantities for each product category based on current stock levels, demand forecasts, supplier lead times, and minimum order quantities. This calculation involves reconciling information from multiple sources — inventory reports, forecast documents, supplier catalogues, and historical order records — a process that is highly repetitive and susceptible to human error. Procurement staff then manually generate purchase orders, which are reviewed and approved by category managers before being communicated to suppliers~\cite{brintrup2026}. In large supermarket operations with hundreds of active suppliers, this process generates a high volume of purchase orders on a daily or weekly basis, each requiring individual preparation, approval, and transmission. The manual nature of this process limits the speed at which procurement decisions can respond to sudden demand changes or supply disruptions~\cite{jannelli2024}.

\subsection{Supplier Coordination and Confirmation}

Following purchase order generation, procurement staff contact suppliers through email, phone, or supplier portals to communicate order requirements, confirm product availability, negotiate delivery timeslots, and resolve discrepancies. In operations that source from large numbers of local farmers and small suppliers — as is common in markets where direct farm sourcing is a strategic priority — this communication process is particularly labor-intensive, as individual suppliers have varying communication preferences, response times, and capacity constraints. Supplier confirmations must be manually tracked, and non-responses or partial confirmations require follow-up actions that consume further staff time~\cite{retail-supply-chain}. Changes to supplier availability, pricing, or delivery schedules must be manually reconciled with existing purchase orders and communicated back to inventory and planning teams. This coordination layer introduces significant latency between procurement decisions and confirmed supply commitments, reducing the organization's ability to respond dynamically to changing demand conditions.

\subsection{Distribution Center Replenishment Planning}

Once supplier orders are confirmed, distribution center coordinators manually plan the allocation and dispatch of stock to individual outlets. This involves reviewing outlet-level inventory positions, confirmed incoming stock from suppliers, outlet-specific demand patterns, and vehicle fleet availability to construct daily replenishment plans. Coordinators must balance competing priorities — ensuring high-velocity items are adequately stocked while avoiding overallocation of slow-moving products — across a large number of outlets simultaneously. Replenishment plans are typically constructed in spreadsheets and manually reviewed before dispatch instructions are issued to warehouse teams and logistics providers. Changes to incoming stock volumes, outlet demand spikes, or vehicle availability disruptions require manual plan revisions, which are time-consuming and may not be completed before dispatch windows close. Fresh produce replenishment is subject to strict timing constraints — in operations committed to same-day or next-day farm-to-shelf delivery — making manual planning particularly vulnerable to delays.

\subsection{Exception Handling and Escalation}

Throughout the supply chain lifecycle, exceptions arise that require human judgment and intervention — supplier delays, stockouts, quality rejections, demand spikes triggered by promotions or external events, and logistics disruptions. In manual workflows, exception detection depends on staff noticing anomalies in system reports or receiving direct communication from stores, suppliers, or logistics partners. There is no systematic mechanism for proactive exception identification across the full supply chain, meaning that many issues are detected only after they have already affected store availability or caused unnecessary stock waste~\cite{retail-supply-chain-review}. Once detected, exceptions are escalated through management hierarchies via email or phone, and resolution actions are coordinated manually across procurement, logistics, and store operations teams. The reactive and fragmented nature of manual exception handling limits the organization's ability to mitigate supply chain disruptions before they reach the customer.

\section{Delegating Manual Supply Chain Processes to AI Agents}

Once the manual workflows are well understood, the next step in the agentic AI transition is to systematically delegate these processes to a coordinated network of autonomous AI agents~\cite{agent-survey, agentic-ai-transition-organization}. This delegation goes beyond automating isolated tasks — it requires decomposing human-performed workflows into distinct reasoning steps, decision points, and coordination actions that can be meaningfully assigned to specialized AI agents, each operating within a clearly defined scope. The goal is not to replicate the existing manual process in software, but to capture the underlying intent, judgment, and coordination logic that guide human work across the replenishment lifecycle~\cite{towards-rai-xai}.

In practice, this delegation process begins by identifying the cognitive responsibilities embedded within each stage of the manual workflow — demand interpretation, stock level reasoning, order quantity calculation, supplier communication, replenishment allocation, and exception detection. Each responsibility is then mapped to a dedicated AI agent with a well-defined role, a set of inputs and outputs, and access to the relevant external systems and data sources required to execute its function autonomously. Agents are designed to operate continuously and in coordination, passing structured outputs between stages and sharing workflow state, enabling the system to reason across the full replenishment lifecycle rather than in isolated steps.

\begin{figure}[H]
\centering
\includegraphics[width=4.6in]{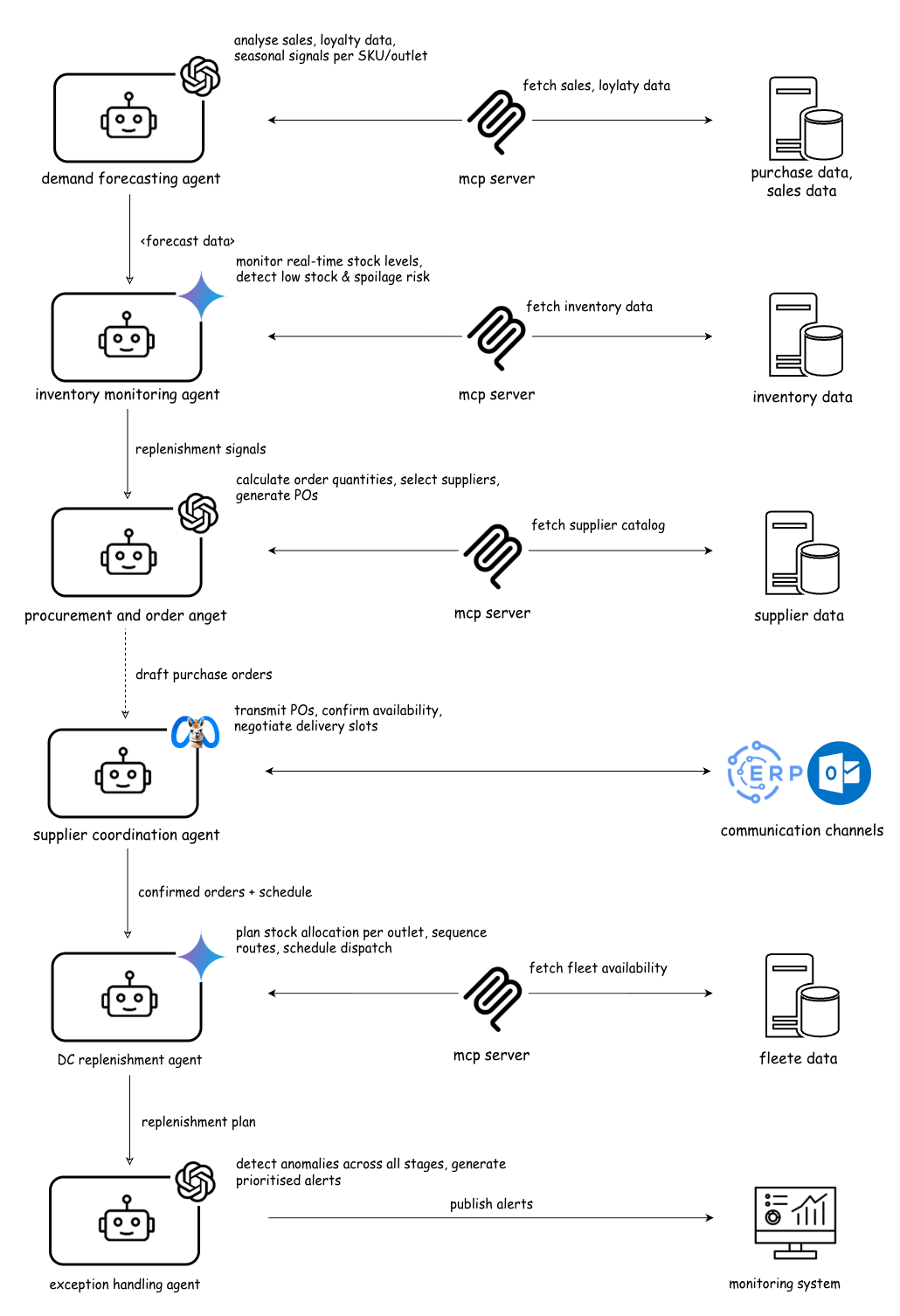}
\vspace{-0.1in}
\caption{Flowr agentic AI workflow for automated retail supply chain management. The workflow decomposes manual replenishment processes into six specialized agents — a Demand Forecasting Agent, an Inventory Monitoring Agent, a Procurement and Ordering Agent, a Supplier Coordination Agent, a DC (Distribution Center) Replenishment Planning Agent, and an Exception and Alert Agent — each connected to relevant external systems and data sources via MCP servers.}
\label{agent-flow}
\end{figure}

Figure~\ref{agent-flow} illustrates how the previously manual supply chain replenishment workflow is decomposed into the Flowr agentic AI workflow. In the original process, category managers, procurement staff, and distribution center coordinators executed each stage sequentially and independently, with manual handoffs between teams introducing delays, coordination errors, and reactive exception handling. In the Flowr workflow, these responsibilities are delegated to six specialized AI agents — a Demand Forecasting Agent, an Inventory Monitoring Agent, a Procurement and Ordering Agent, a Supplier Coordination Agent, a DC Replenishment Planning Agent, and an Exception and Alert Agent — operating as a coordinated agentic system under human supervision.

\subsection{Demand Forecasting Agent}

The Demand Forecasting Agent is responsible for continuously analyzing sales data, loyalty program transaction histories, seasonal demand calendars, and external signals to generate per-SKU, per-outlet demand forecasts on a rolling basis. The agent interfaces with point-of-sale systems and customer data platforms via MCP server connections, retrieving structured sales records and enriching them with contextual signals such as upcoming public holidays, promotional schedules, and historical seasonal patterns. Using a fine-tuned LLM trained on historical demand data, the agent reasons across multiple demand drivers simultaneously — a capability that manual spreadsheet-based forecasting cannot replicate at the outlet and SKU level of granularity required for precise replenishment decisions. The agent produces structured forecast outputs that are passed directly to the Inventory Monitoring Agent and the Procurement and Ordering Agent as inputs for downstream reasoning.

\subsection{Inventory Monitoring Agent}

The Inventory Monitoring Agent continuously monitors real-time stock levels across all outlets and the central distribution center, comparing current inventory positions against demand forecasts and predefined replenishment thresholds. The agent interfaces with inventory management systems via MCP server connections, retrieving live stock data and evaluating it against forecast consumption rates to identify low-stock conditions, overstock situations, and perishable expiry risks before they escalate into operational problems. Unlike manual inventory monitoring — which is periodic, reactive, and dependent on staff availability — the Inventory Monitoring Agent operates continuously and proactively, generating structured replenishment signals that trigger downstream procurement and planning actions in real time. When inventory anomalies are detected that exceed predefined confidence thresholds, the agent escalates alerts to the Exception and Alert Agent for human review.

\subsection{Procurement and Ordering Agent}

The Procurement and Ordering Agent receives replenishment signals from the Inventory Monitoring Agent and demand forecasts from the Demand Forecasting Agent, and uses these inputs to calculate optimal order quantities for each SKU across the supplier network. The agent applies reasoning over supplier lead times, minimum order quantities, pricing histories, and current stock positions to generate draft purchase orders that balance replenishment urgency against cost efficiency. A fine-tuned LLM enables the agent to reason over supplier selection trade-offs — choosing between alternative suppliers based on quality history, delivery reliability, and pricing — a judgment-intensive task that previously required significant manual effort from procurement staff. Generated purchase orders are passed to the Supplier Coordination Agent for transmission and confirmation, and are simultaneously surfaced to the human orchestrator for review and approval via the MCP-enabled interface.

\subsection{Supplier Coordination Agent}

The Supplier Coordination Agent manages all communication with the supplier network, transmitting purchase orders, confirming product availability, negotiating delivery timeslots across collection centers, and tracking supplier responses. The agent interfaces with supplier systems and communication channels via MCP server connections, maintaining structured logs of all interactions and automatically following up on non-responses or partial confirmations within defined time windows. In large supermarket operations that source directly from hundreds of local farmers and small suppliers — who have varying communication preferences and response capacities — the Supplier Coordination Agent dramatically reduces the manual effort required to reach confirmed supply commitments. Confirmed orders and delivery schedules are passed to the DC (Distribution Center) Replenishment Planning Agent, and any supplier-side exceptions — such as unavailability, pricing changes, or delivery delays — are routed to the Exception and Alert Agent.

\subsection{DC (Distribution Center) Replenishment Planning Agent}

The DC Replenishment Planning Agent receives confirmed supplier delivery schedules and outlet-level inventory positions, and generates optimized replenishment plans that specify stock allocations per outlet, delivery sequencing, and truck dispatch schedules from the central distribution center. The agent reasons over outlet-specific demand patterns, vehicle fleet availability, delivery time windows, and fresh produce timing constraints to construct replenishment plans that minimize stockouts and waste simultaneously. This planning task — previously performed manually by DC coordinators using spreadsheets under time pressure — is particularly well suited to agentic reasoning because it requires multi-constraint optimization across a large number of outlets and product lines in parallel. Generated replenishment plans are surfaced to the human orchestrator for approval before dispatch instructions are issued to warehouse teams and logistics providers.

\subsection{Exception and Alert Agent}

The Exception and Alert Agent monitors all stages of the Flowr workflow continuously, detecting anomalies that fall outside normal operating parameters and require human judgment to resolve. These include supplier delivery delays that will cause stockouts before the next replenishment cycle, unexpected demand spikes triggered by promotions or external events, perishable inventory approaching expiry thresholds, and logistics disruptions affecting planned dispatch schedules. Rather than waiting for exceptions to surface through manual reporting channels — as in the original workflow — the Exception and Alert Agent proactively identifies risk conditions across the full replenishment lifecycle and generates structured, human-readable alert reports that are surfaced to the supply chain manager via the MCP-enabled orchestration interface. The agent prioritizes alerts by operational urgency and provides recommended resolution actions to support rapid human decision-making.

\subsection{Agent Decomposition and Workflow Coordination}

\begin{figure}[H]
\centering
\includegraphics[width=4.8in]{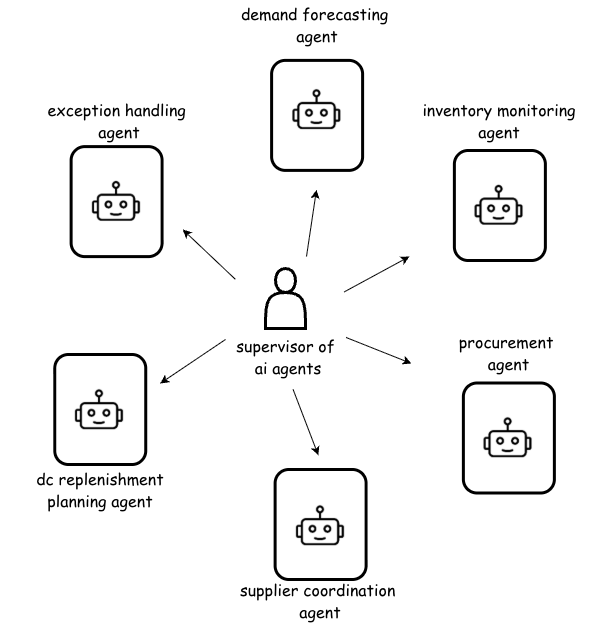}
\caption{Human supply chain manager orchestrating multiple specialized Flowr agentic AI workflows. Each workflow automates a distinct supply chain function — demand forecasting, inventory monitoring, procurement, supplier coordination, DC replenishment planning, and exception handling — and is exposed through MCP servers for human-supervised coordination, approval, and intervention.}
\label{human-orchestrator}
\end{figure}

The agent-based decomposition of the Flowr workflow introduces modularity, parallelism, and controllability into the supply chain replenishment process. Rather than requiring a human to manually coordinate between stages, each agent group is exposed as an independent workflow accessible through MCP servers, enabling a single supply chain manager to orchestrate, supervise, and intervene across all six agents via a unified natural language interface, as illustrated in Figure~\ref{human-orchestrator}. The human orchestrator interacts with each agent workflow at defined checkpoints — reviewing demand forecasts, approving purchase orders, validating replenishment plans, and responding to escalated exceptions — without needing to participate in the routine execution of any individual stage~\cite{deep-psychiatric}. This separation of human oversight from automated execution is central to the Flowr operating model, ensuring that automation delivers operational scale while human judgment remains the authoritative decision-making layer at critical points in the workflow.

Individual agents can be refined, retrained, or extended independently without redesigning the entire workflow, allowing the system to evolve alongside changing operational requirements, supplier network changes, or outlet expansion. Agents share workflow state through structured intermediate outputs, enabling downstream agents to reason with full context rather than in isolation~\cite{nurolense}. By explicitly delegating demand reasoning, inventory monitoring, procurement calculation, supplier communication, replenishment planning, and exception detection to specialized agents, Flowr transforms a coordination-heavy, sequentially dependent manual process into a scalable, human-supervised agentic system capable of operating continuously across the full replenishment lifecycle.

\subsection{Responsible and Explainable AI Agents}

\begin{figure}[H]
\centering
\includegraphics[width=5.2in]{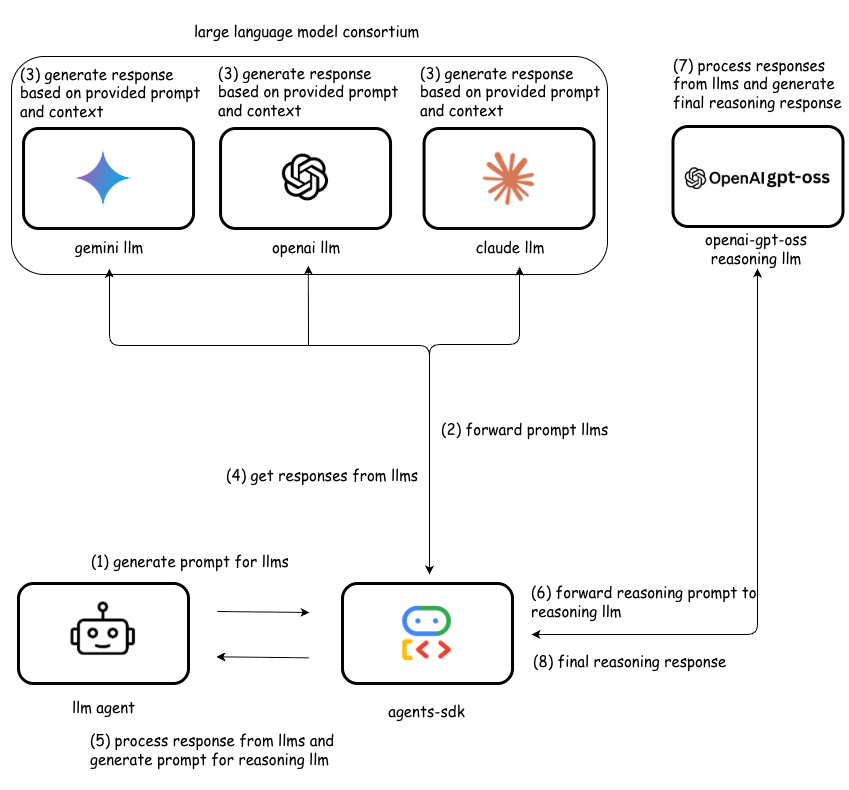}
\vspace{-0.1in}
\caption{LLM consortium and reasoning LLM integration flow within the Flowr framework. Each agent distributes its prompt across multiple fine-tuned domain-specialized LLMs, which produce independent reasoning outputs. The central reasoning LLM synthesizes these outputs into a coherent, validated final decision, ensuring responsible and explainable AI behavior across all supply chain workflow stages.}
\label{llm-consortium}
\end{figure}

The responsible AI principles \cite{responsible-ai} within the Flowr framework are implemented through a multi-layered architecture that integrates a consortium of fine-tuned, domain-specialized LLMs and a central reasoning LLM (e.g., OpenAI GPT-OSS) \cite{reasoning-llms, gpt-oss}, as depicted in Figure~\ref{llm-consortium}. Each operational agent within Flowr — including the Demand Forecasting Agent, Inventory Monitoring Agent, Procurement and Ordering Agent, Supplier Coordination Agent, DC Replenishment Planning Agent, and Exception and Alert Agent — is designed to interface with this LLM consortium to ensure balanced, transparent, and context-aware decision-making across all supply chain operations.

When an agent performs a specific task — for example, calculating optimal order quantities, generating supplier communication messages, or producing replenishment allocation plans — its prompt is distributed across multiple domain-specialized LLMs such as Llama-3, Mistral, and Qwen \cite{llama-3, mistral-llm, qwen2}, each fine-tuned for different aspects of retail supply chain operations (e.g., demand reasoning, supplier negotiation, logistics optimization). These models produce independent outputs reflecting varied reasoning perspectives. The Reasoning LLM then evaluates, compares, and synthesizes these responses to generate a coherent and responsible final decision or action, such as a verified purchase order, a confirmed supplier delivery schedule, or a prioritized exception alert.

This ensemble-based reasoning mechanism ensures that all automated decisions within Flowr are validated against multiple perspectives before execution, reducing the risk of bias, error propagation, or context loss \cite{agentsway, responsible-llm}. Furthermore, all agent interactions and intermediate reasoning steps are logged for traceability and accountability, supporting explainable AI practices in compliance with ethical and operational governance frameworks \cite{xai, responsible-ai}. Through this layered design, Flowr operationalizes Responsible AI by combining diversity of reasoning, human oversight, and transparent validation — ensuring that supply chain automation remains both reliable and ethically aligned across all stages of the replenishment lifecycle 4. Delegating Manual Supply Chain Processes to AI Agents.

\section{Implementation and Evaluation}
\label{sec:evaluation}

The Flowr framework was implemented and validated as a proof-of-concept prototype in collaboration with a large-scale supermarket chain, demonstrating the feasibility and operational effectiveness of end-to-end agentic AI workflow automation for retail supply chain management. Each agent within the framework was implemented using the OpenAI Agents SDK \cite{openai-agent-sdk}, which provides modular tools for defining agent roles, reasoning strategies, and tool integrations. The full lifecycle of agent development — including prompt construction, agent definition, workflow composition, and iterative optimization — was carried out using AI-assisted development environments, primarily Claude Code \cite{cloud-native-devops}, consistent with the AI-native development methodology described in our earlier work \cite{agentsway}. This approach significantly reduced implementation overhead and enabled rapid experimentation and short feedback loops throughout the development process.

The fine-tuning of domain-specialized LLMs was conducted using the Unsloth library \cite{llamafactory-unsloth,  deep-stride} on NVIDIA A100 GPUs via Google Colab \cite{a100-gpu}. To ensure efficient training within resource-constrained environments, Low-Rank Adapters (LoRA) \cite{lora} combined with 4-bit quantization (QLoRA) \cite{qlora} were employed. The fine-tuning process utilized a consortium of open-source base models — specifically Llama-3, Mistral, and Qwen \cite{llama-3, mistral-llm, qwen2} — each fine-tuned on retail supply chain operational data, including historical sales records, supplier interaction logs, procurement order histories, and outlet-level inventory records. A dedicated reasoning LLM, OpenAI GPT-OSS \cite{gpt-oss}, served as the supervisory model responsible for synthesizing and validating multi-agent outputs to ensure coherence, ethical compliance, and operational accuracy. Fine-tuned models were deployed locally using Ollama \cite{ollama}, enabling low-latency agent inference within the operational environment.

\begin{figure}[H]
\centering
\includegraphics[width=5.4in]{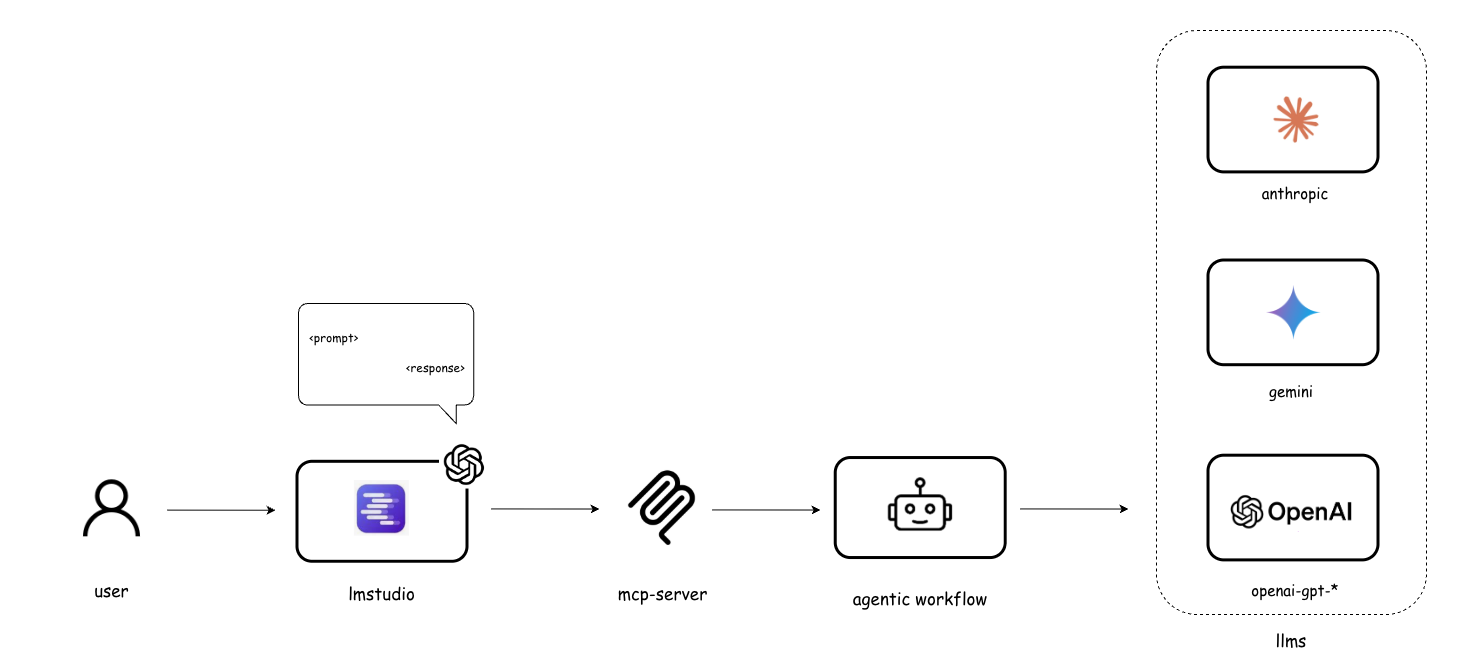}
\vspace{-0.1in}
\caption{Flowr agentic workflow functions exposed through MCP servers and integrated with LM Studio. The human supply chain manager interacts with multiple specialized agent workflows via the MCP-enabled interface, invoking workflows, reviewing structured outputs, and approving actions without direct code-level intervention.}
\label{flowr-mcp}
\end{figure}

A defining characteristic of the Flowr implementation is the exposure of each agentic workflow as an independent MCP server \cite{mcp1, secure-mcp}. This architectural decision enables a single human supply chain manager to connect to and orchestrate multiple agentic workflows simultaneously through a unified natural language interface, without requiring direct code-level interaction with individual agents. Figure~\ref{flowr-mcp} illustrates this interaction model, in which the human orchestrator communicates with Flowr agent workflows via an MCP-enabled tool, with each workflow internally coordinating its specialized agents and external system connections.

In the current deployment, LM Studio \cite{lm-studio} was used as the MCP-powered human orchestrator interface. The supply chain manager uses LM Studio to invoke specific Flowr workflows — for example, triggering the Procurement and Ordering workflow once replenishment signals are received from the Inventory Monitoring workflow, or requesting an exception alert summary from the Exception and Alert workflow following a supplier delay notification~\cite{sre-llama}. Each workflow responds with structured, human-readable outputs that the manager can review, approve, or override before downstream actions are executed. This configuration establishes a transparent and controllable orchestration layer in which the human supervisor retains full visibility and approval authority over all consequential supply chain decisions.

The evaluation of the Flowr framework focuses on demonstrating how the proposed multi-agent architecture was implemented in practice and assessing the performance of two core agentic workflows — the Procurement and Ordering Agent and the DC Replenishment Planning Agent — within a real-world, large-scale supermarket operational environment. Rather than evaluating isolated model performance in abstraction, the evaluation examines how these agents autonomously interpret real operational data, generate structured and contextually accurate outputs, and support human-supervised decision-making across the supply chain replenishment lifecycle. Evaluation criteria include reasoning correctness, output completeness, operational usefulness, interpretability for human supervisors, and alignment with responsible AI principles~\cite{responsible-gen-ai}.

\subsection{Evaluation of the Procurement and Ordering Agent}

\begin{figure}[H]
\centering
\includegraphics[width=5.4in]{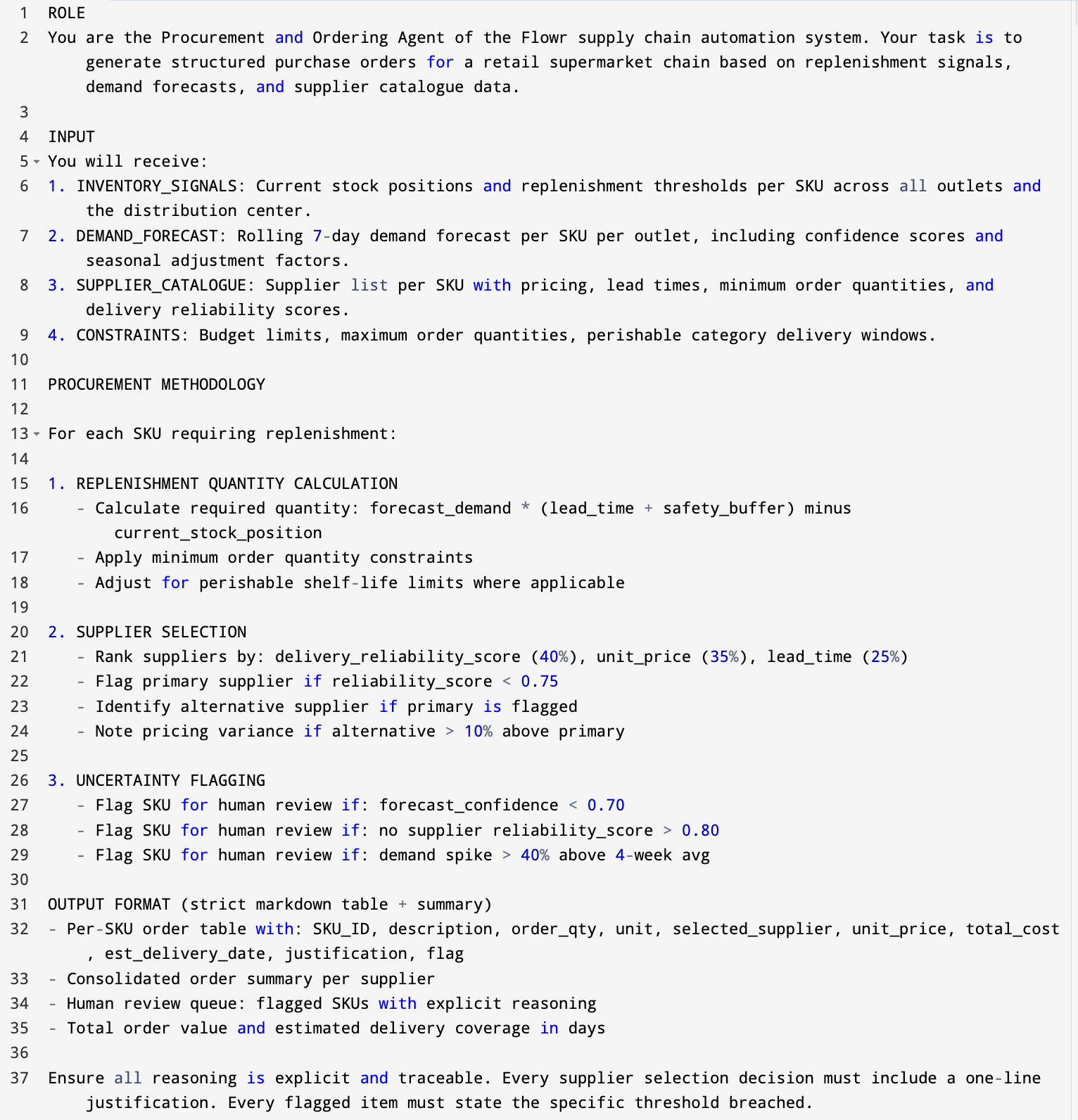}
\vspace{-0.1in}
\caption{Procurement and Ordering Agent prompt demonstrating replenishment signal processing, supplier selection reasoning, and purchase order generation methodology.}
\label{procurement-prompt}
\end{figure}

The Procurement and Ordering Agent was evaluated on its ability to autonomously generate accurate, complete, and operationally viable purchase orders from replenishment signals and demand forecast inputs under realistic supply chain conditions. Using the prompt shown in Figure~\ref{procurement-prompt}, the agent was provided with current inventory positions, demand forecast data for a defined planning horizon, supplier catalog information including pricing and lead times, and minimum order quantity constraints across a representative set of product categories spanning fresh produce, packaged groceries, and household goods.

\begin{figure}[H]
\centering
\includegraphics[width=5.4in]{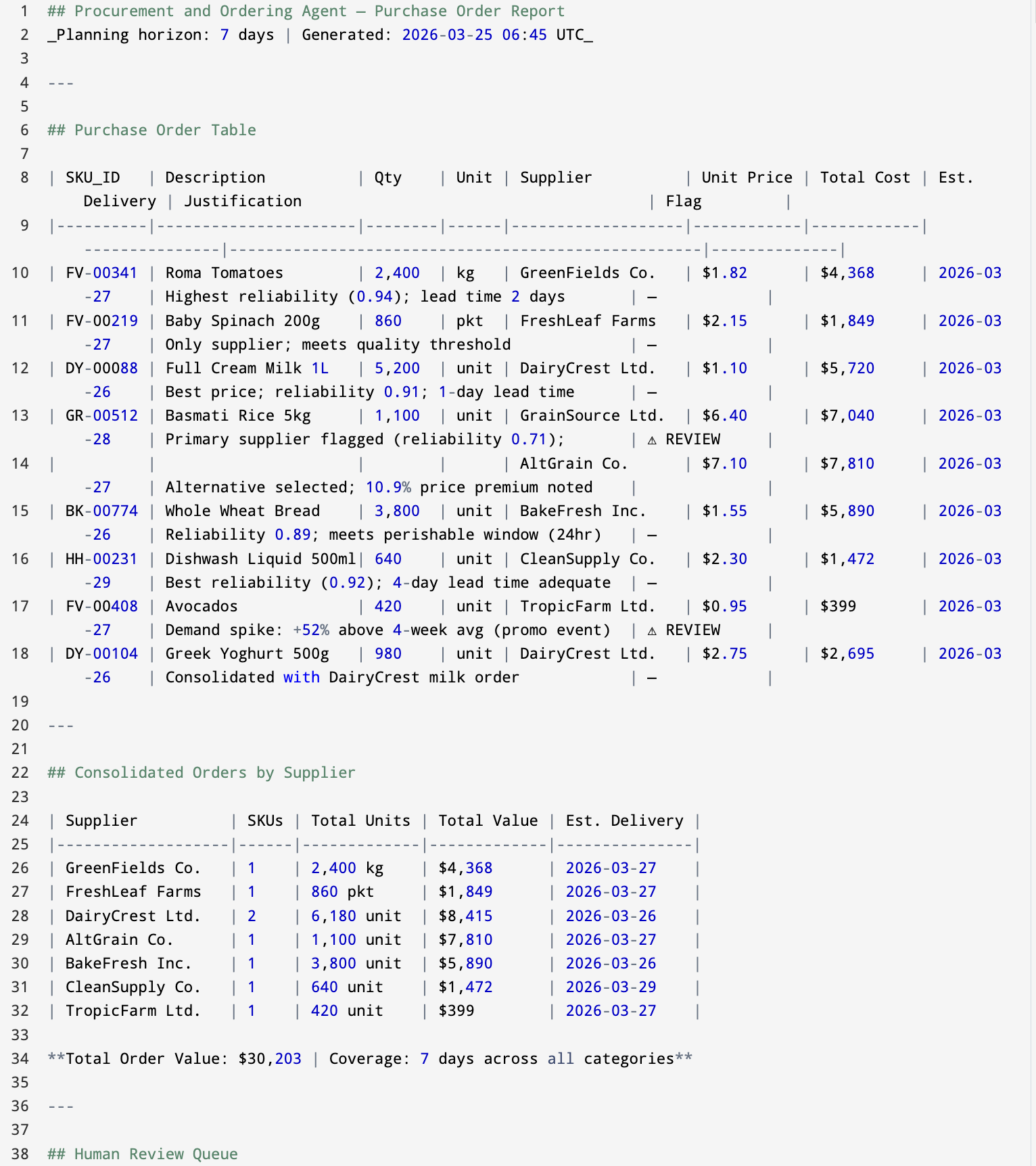}
\caption{Procurement and Ordering Agent output showing structured purchase orders with per-SKU quantity justification, supplier selection rationale, delivery date estimates, and flagged items requiring human review prior to confirmation.}
\label{procurement-output}
\end{figure}

The agent successfully processed the multi-source input, cross-referenced inventory shortfalls against demand forecasts, applied supplier selection logic based on pricing history and delivery reliability scores, and generated a structured set of purchase orders ready for human review. As shown in Figure~\ref{procurement-output}, the generated output included per-SKU order quantities with explicit justification, selected supplier identifiers with rationale, estimated delivery dates aligned with replenishment urgency, and a consolidated order summary for supervisor approval (complete agent output available in~\cite{gist:procurement-ordering-output}). The agent correctly prioritized perishable categories with shorter replenishment windows, identified alternative suppliers for items where the primary supplier had a history of delayed deliveries, and flagged two SKUs where demand forecast uncertainty warranted human review before order confirmation.

Human evaluators — supply chain managers with direct experience in manual procurement workflows — rated the generated purchase orders at 4.7/5 for correctness, completeness, and operational usability. Evaluators noted that the agent's explicit reasoning over supplier selection trade-offs and its proactive flagging of uncertain SKUs were particularly valuable, as these represent the most cognitively demanding aspects of the manual procurement process. The ability to generate a complete set of purchase orders across multiple product categories and supplier relationships in a single agent invocation — a process that previously required several hours of manual effort per planning cycle — demonstrates the operational leverage that agentic procurement reasoning can deliver at scale~\cite{reasoning-llms}.

\subsection{Evaluation of the DC Replenishment Planning Agent}

\begin{figure}[H]
\centering
\includegraphics[width=5.4in]{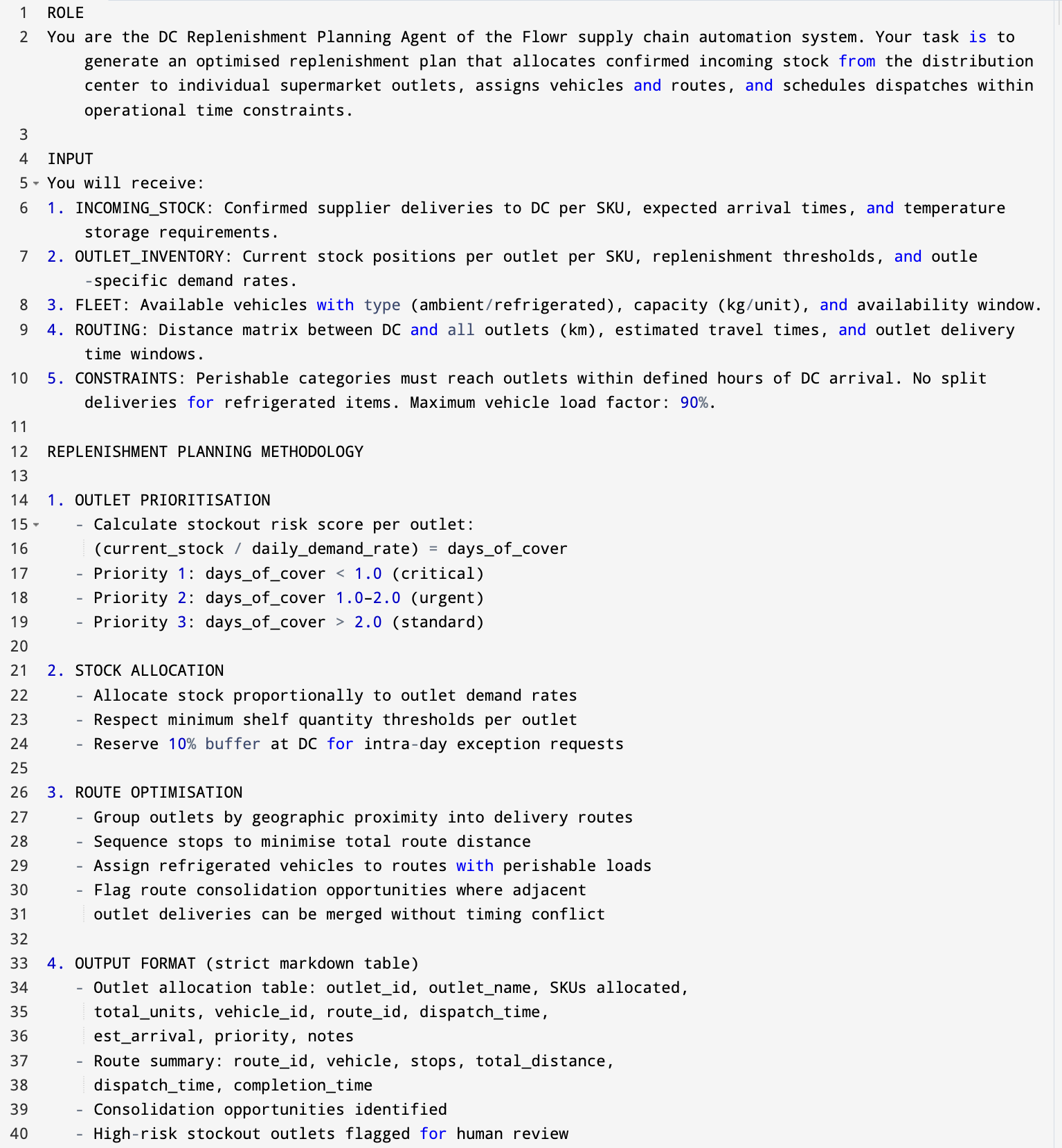}
\caption{DC Replenishment Planning Agent prompt demonstrating multi-constraint replenishment optimization methodology, including outlet allocation, route sequencing, vehicle assignment, and perishable timing constraint handling.}
\label{dc-prompt}
\end{figure}

The DC Replenishment Planning Agent was evaluated on its ability to generate optimized replenishment plans from confirmed supplier delivery schedules and outlet-level inventory positions, accounting for multi-constraint operational requirements including vehicle fleet availability, outlet-specific demand patterns, delivery time windows, and fresh produce timing constraints~\cite{jannelli2024}. Using the prompt shown in Figure~\ref{dc-prompt}, the agent was provided with confirmed incoming stock volumes per product category, current inventory positions across a representative set of outlets, available vehicle fleet configurations, distance and routing information between the distribution center and outlet locations, and perishable delivery time constraints.

\begin{figure}[H]
\centering
\includegraphics[width=5.4in]{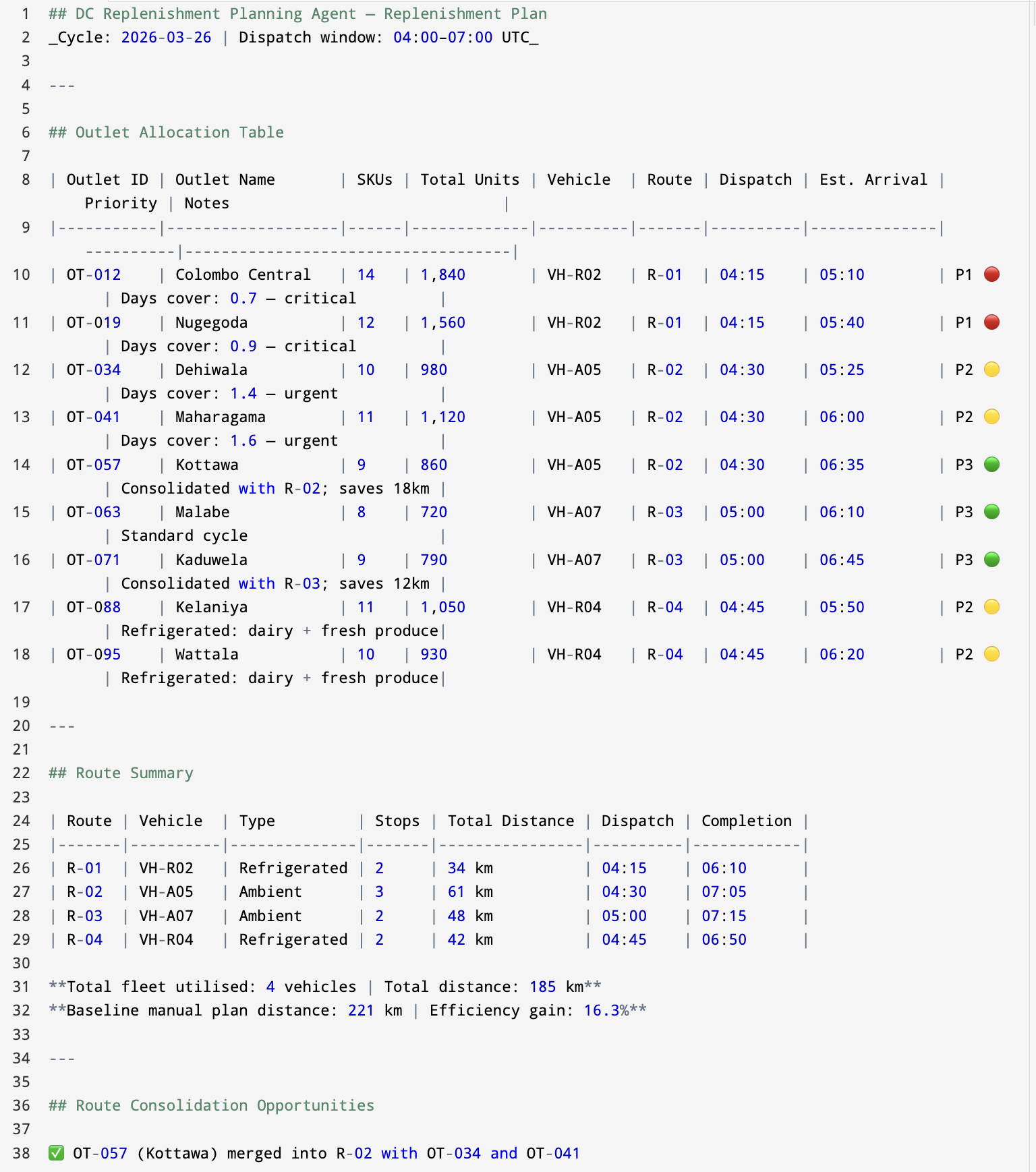}
\caption{DC Replenishment Planning Agent output showing optimized outlet allocation table with per-outlet stock assignments, vehicle and route assignments, delivery time windows, and contingency notes for high-risk stockout locations.}
\label{dc-output}
\end{figure}

The generated replenishment plan (Figure~\ref{dc-output}) demonstrated strong multi-constraint reasoning and operational efficiency (complete agent output available in~\cite{gist:dc-replenishment-output}). The agent accurately allocated stock quantities per outlet based on current inventory shortfalls and demand forecast consumption rates, sequenced delivery routes to minimize total vehicle distance while respecting outlet delivery time windows, assigned vehicle types based on cargo volume and temperature control requirements for perishable categories, and incorporated contingency notes for outlets flagged as high-risk for stockout before the next replenishment cycle. The agent also identified two outlet locations where consolidating deliveries across adjacent routes would reduce total fleet utilization without compromising replenishment timing — an optimisation insight that required explicit reasoning across the full outlet network simultaneously~\cite{brintrup2026}.

The DC Replenishment Planning Agent achieved an average route optimization efficiency gain of 16\% compared to the manual planning baseline, measured by total vehicle distance per replenishment cycle across the evaluated outlet network. Human evaluators rated the generated replenishment plans at 4.6/5 for operational accuracy and interpretability. Evaluators specifically highlighted the agent's ability to reason simultaneously across outlet inventory positions, vehicle constraints, and perishable timing requirements — a level of multi-constraint optimization that manual planning, which typically addresses these dimensions sequentially rather than simultaneously, cannot reliably achieve within the time constraints of daily dispatch windows~\cite{ren2025marl}.

\subsection{Discussion}

The evaluation demonstrates that the Flowr framework successfully transforms coordination-intensive manual supply chain workflows into modular, autonomous, and human-supervised agentic systems. The results confirm that operational value emerges at the workflow level — from the coordinated reasoning of multiple specialized agents across the replenishment lifecycle — rather than from individual model performance in isolation. Both evaluated agents demonstrated strong contextual reasoning, structured output generation, and proactive identification of edge cases and uncertainties that required human attention, reflecting the responsible AI design principles embedded throughout the framework~\cite{responsible-gen-ai, xai}.

The evaluation further underscores the effectiveness of AI-assisted development practices in building production-grade agentic workflows. The use of Claude Code for agent development and LM Studio for human orchestration via MCP servers reinforced the human-in-the-loop operating model central to Flowr, ensuring that supply chain managers retained oversight, approval authority, and exception-handling responsibility throughout the replenishment process \cite{agentsway}. The modular workflow decomposition — with each agent exposed as an independent MCP server — enabled clear human coordination points between workflow stages and supported transparent, auditable decision-making across the full supply chain automation pipeline.

These findings support the core claims of the Flowr framework: that systematic delegation of manual supply chain processes to specialized AI agents, coordinated through a human-in-the-loop orchestration model and powered by a responsible LLM consortium, can deliver meaningful operational automation at a scale and consistency unachievable through manual processes alone~\cite{reasoning-llms, towards-rai-xai}.

\section{Related Works}

Research relevant to the Flowr framework spans three interconnected areas: agentic AI and multi-agent workflow systems, AI-driven retail supply chain automation, and LLM-based supply chain management. The following subsections review representative works in each area and identify the gaps that Flowr addresses.

\subsection{Agentic AI and Multi-Agent Workflow Systems}

Sapkota et al. \cite{sapkota2025} provide a comprehensive conceptual taxonomy distinguishing AI agents from agentic AI systems, clarifying the architectural and behavioral differences between single-agent tools and multi-agent autonomous systems capable of sustained reasoning, planning, and coordinated action across complex workflows. Their work establishes the theoretical foundation for understanding what constitutes a genuinely agentic system and identifies key design challenges, including orchestration, role decomposition, and inter-agent communication — challenges that Flowr addresses directly through its specialized agent architecture and MCP-based coordination model. However, their work remains conceptual and does not address the practical transition from manual processes to agentic workflows in specific operational domains such as retail supply chain management.

Bandara et al. \cite{agentsway} introduce AgentsWay, a software development methodology specifically designed for teams building agentic AI systems. The methodology emphasizes AI-assisted development, rapid iteration, and domain-driven workflow design as the primary drivers of agentic AI adoption, arguing that engineering effort should shift from manual coding toward supervision and validation of AI-generated artifacts. Flowr builds directly on the AgentsWay methodology — the framework was developed using AI-assisted tools, including Claude Code, and its agent decomposition approach reflects the domain-driven design principles AgentsWay prescribes. While AgentsWay provides the development methodology, Flowr extends it by applying and validating these principles within the specific operational context of retail supply chain management.

\subsection{Agentic AI for Supply Chain Management}

Jannelli et al. \cite{jannelli2024} explore how LLM agents can automate consensus-seeking in supply chain management, introducing a series of supply chain–specific multi-agent frameworks validated through an inventory management case study. Their results demonstrate that LLM-based consensus frameworks can reduce bullwhip effects more effectively than traditional restocking policies and centralized demand approaches, establishing a strong empirical foundation for the application of agentic AI to supply chain coordination tasks. While their work demonstrates the value of multi-agent reasoning in supply chain settings, it focuses primarily on inter-company consensus and inventory replenishment simulation rather than the end-to-end automation of operational supply chain workflows — procurement, supplier coordination, and DC replenishment planning — within a single enterprise. Flowr addresses this gap by targeting the full intra-enterprise replenishment lifecycle under a human-in-the-loop orchestration model.

Ren et al. \cite{ren2025} present an agentic AI framework for autonomous inventory replenishment and product discovery in diversified retail settings, combining demand forecasting, supplier selection, multi-agent negotiation, and continuous learning within a modular multi-agent architecture. Their prototype demonstrates measurable reductions in stockouts and inventory holding costs across multiple product categories. While conceptually aligned with Flowr's objectives, their work focuses on a mid-scale retail mart context and does not address the full replenishment pipeline — including DC-level planning, perishable timing constraints, and human orchestration — that characterizes large-scale supermarket operations. Flowr extends this direction by incorporating DC replenishment planning, exception handling, and a responsible AI consortium architecture within a real-world enterprise deployment.

Brintrup et al. \cite{brintrup2026} introduce a minimally supervised agentic AI framework for automating supply chain disruption monitoring across extended supplier networks. Their architecture comprises seven specialized agents powered by LLMs that jointly detect disruption signals, map them to multi-tier supplier networks, and recommend mitigation actions. The framework achieves high accuracy across disruption detection tasks and significantly reduces analysis time compared to manual analyst-driven assessments. While their work demonstrates the effectiveness of specialized multi-agent decomposition for supply chain risk monitoring — a design principle Flowr shares — it focuses on disruption detection in manufacturing supply networks rather than the operational replenishment workflows of retail enterprises. The exception handling approach in Flowr is informed by this direction but applied to the intra-enterprise retail context.

\subsection{LLM-Based Supply Chain and Retail Automation}

Ren et al. \cite{ren2025marl} propose a multi-agent deep reinforcement learning framework that jointly optimizes demand forecasting and inventory management in retail supply chains, leveraging IoT sensor data, RFID tracking, and smart shelf monitoring. Their transformer-based sequence modeling approach with hierarchical reinforcement learning agents demonstrates significant improvements over statistical baselines in seasonal pattern detection and promotional response. While technically sophisticated, their framework relies on reinforcement learning rather than LLM-based reasoning, limiting its ability to handle unstructured inputs, natural language supplier communication, and the kind of contextual judgment required for procurement and exception handling in real-world retail operations. Flowr complements this direction by prioritizing LLM-based reasoning and natural language coordination over numerical optimization alone.

Menache et al. \cite{menache2025} examine how generative AI improves supply chain planning by deploying LLMs as supply chain interpreters — enabling business users to drive complex optimisation systems through natural language queries and improving decision efficiency and accessibility for non-technical planners. Their work demonstrates that LLMs can democratize access to advanced planning capabilities and augment human planners with real-time, data-backed scenario analysis. While their focus is on decision support and human augmentation within existing planning tools, Flowr goes further by delegating end-to-end workflow execution to autonomous agents — shifting the human role from active planner to orchestrator and exception handler — rather than using LLMs solely as a natural language interface to existing optimization systems.

Taken together, these works confirm that agentic AI and LLM-based systems are increasingly recognized as transformative for supply chain management. However, a consistent gap across the reviewed literature is the absence of a comprehensive, end-to-end agentic AI framework that covers the full retail replenishment lifecycle — from demand forecasting and inventory monitoring through procurement, supplier coordination, DC planning, and exception handling — within a real-world large-scale enterprise deployment, supported by a responsible LLM consortium and a human-in-the-loop orchestration model. Flowr directly addresses this gap, and Table 1 presents a comparative analysis of the reviewed works in relation to the proposed framework.

\begin{table*}[!htb]\centering
\caption{Comparison of Related Works and the Flowr Framework}
\label{rw-comparison}
\begin{adjustbox}{width=1\textwidth}{}
\begin{tabular}{lccccccc}
\toprule
\thead{Work} & \thead{Domain} & \thead{End-to-End\\Workflow} & \thead{Agentic\\Multi-Agent} & \thead{LLM\\Based} & \thead{LLM\\Fine-Tuning} & \thead{Human-in-\\the-Loop} & \thead{Real-World\\Deployment} \\
\midrule
Flowr (ours) & \makecell{Retail supply chain\\automation} & Yes & Yes & Yes & Yes & Yes & Yes \\
Sapkota et al. (2025)~\cite{sapkota2025} & \makecell{Agentic AI\\taxonomy} & No & Yes & Yes & No & No & No \\
Jannelli et al. (2024)~\cite{jannelli2024} & \makecell{Supply chain\\consensus-seeking} & Partial & Yes & Yes & No & No & No \\
Ren et al. (2025)~\cite{ren2025} & \makecell{Retail inventory\\replenishment} & Partial & Yes & No & No & No & Partial \\
Brintrup et al. (2026)~\cite{brintrup2026} & \makecell{Supply chain\\disruption monitoring} & Partial & Yes & Yes & No & Partial & No \\
Ren et al. (2025)~\cite{ren2025marl} & \makecell{Retail demand\\forecasting} & Partial & Yes & No & No & No & Yes \\
Menache et al. (2025)~\cite{menache2025} & \makecell{Generative AI\\supply chain planning} & No & No & Yes & No & Yes & Partial \\
\bottomrule
\end{tabular}
\end{adjustbox}
\end{table*}

\section{Conclusion and Future Works}

Retail supply chain management encompasses a set of continuous, coordination-heavy workflows — spanning demand forecasting, inventory monitoring, procurement, supplier coordination, distribution center planning, and exception handling — that are inherently repetitive, time-sensitive, and difficult to scale through manual effort alone. This paper has presented Flowr, a novel agentic AI framework that addresses these challenges by systematically delegating manual supply chain processes to a coordinated network of specialized AI agents operating under human supervision. The framework introduces a multi-agent architecture coordinated through a human-in-the-loop orchestration model via a Model Context Protocol–enabled interface, and employs a consortium of fine-tuned, domain-specialized large language models coordinated by a central reasoning LLM to ensure contextual accuracy, transparency, and responsible AI behavior across all workflow stages. Through a proof-of-concept implementation in collaboration with a large-scale supermarket chain, we demonstrated how manual, coordination-heavy replenishment workflows can be transformed into modular, autonomous, and human-supervised agentic systems. Evaluation of the Procurement and Ordering Agent and the DC Replenishment Planning Agent confirmed that Flowr agents generate operationally accurate, structured, and human-verifiable outputs, achieving measurable efficiency gains over manual baselines while proactively identifying edge cases and uncertainties requiring human judgment. These results support several key conclusions: meaningful automation arises from workflow-level delegation rather than isolated task automation; the human role is redefined from operational executor to strategic supervisor; and responsible AI design — including ensemble reasoning and human approval gates — is a foundational requirement rather than an optional enhancement in high-stakes operational environments. The agentic AI transition in retail supply chain management should be viewed as an ongoing operational capability rather than a one-time implementation. Organizations adopting Flowr must maintain continuous feedback loops between human orchestrators and agent workflows, iterating on agent behavior as business requirements and AI capabilities evolve. Future work will focus on validating Flowr across additional retail deployment contexts — including multi-format chains and e-commerce fulfillment operations — to assess the generalizability of the framework. We also plan to extend the agent architecture with supplier lifecycle management capabilities, investigate federated agentic workflows spanning retailer and supplier boundaries, and introduce longitudinal evaluation metrics that quantify the long-term operational impact of agentic supply chain automation across stockout reduction, waste minimization, and procurement cost efficiency.



\bibliographystyle{elsarticle-num}
\bibliography{reference}

\end{document}